\documentclass[letterpaper]{article} % DO NOT CHANGE THIS
\usepackage{aaai23}  % DO NOT CHANGE THIS
\usepackage{times}  % DO NOT CHANGE THIS
\usepackage{helvet}  % DO NOT CHANGE THIS
\usepackage{courier}  % DO NOT CHANGE THIS
\usepackage[hyphens]{url}  % DO NOT CHANGE THIS
\usepackage{graphicx} % DO NOT CHANGE THIS
\usepackage{hyperref}
\usepackage{natbib}  % DO NOT CHANGE THIS AND DO NOT ADD ANY OPTIONS TO IT
\usepackage{caption} % DO NOT CHANGE THIS AND DO NOT ADD ANY OPTIONS TO IT
\DeclareCaptionStyle{ruled}{labelfont=normalfont,labelsep=colon,strut=off} % DO NOT CHANGE THIS
\usepackage[T1]{fontenc}

\usepackage{dsfont}
\usepackage{subfig}
\usepackage{amsmath}
\usepackage{multirow}
\usepackage{textcomp}
\usepackage{tabularx}
\usepackage{enumerate}
\usepackage{nameref}
\usepackage{graphicx}
\usepackage{appendix}
\usepackage{wrapfig}
\usepackage{makecell}

\usepackage{algorithm}
\usepackage{algpseudocode}
\usepackage{mathtools}

\usepackage{comment}

\makeatletter
\def\RS#1{\zap@space#1 \@empty}
\makeatother

\usepackage{url}            % simple URL typesetting
\usepackage{booktabs}       % professional-quality tables
\usepackage{amsfonts}       % blackboard math symbols
\usepackage{nicefrac}       % compact symbols for 1/2, etc.
\usepackage{microtype}      % microtypography
\usepackage{xcolor}         % colors
\usepackage{ulem}

\title{Spatial-Temporal Graph Convolutional Gated Recurrent Network for Traffic Forecasting}

 \author{%
  \href{mailto:2209557029@qq.com}{Le Zhao}, \href{mailto:chenmc@smail.nju.edu.cn}{Mingcai Chen}, \href{mailto:duyuntao@smail.nju.edu.cn}{Yuntao Du}, \href{mailto:hyyang@smail.nju.edu.cn}{Haiyang Yang}, \href{mailto:chjwang@nju.edu.cn}{Chongjun Wang}\\%\thanks{Corresponding authors}\\
}
\affiliations{
   State Key Laboratory for Novel Software Technology at Nanjing University\\
   Nanjing University, Nanjing 210023, China\\
  {2209557029}@qq.com,\\
  \{chenmc, duyuntao, hyyang\}@smail.nju.edu.cn, \\
  {chjwang}@nju.edu.cn \\
}
\begin{document}
\maketitle
\begin{abstract}
    As an important part of intelligent transportation systems, traffic forecasting has attracted tremendous attention from academia and industry. 
    Despite a lot of methods being proposed for traffic forecasting, it is still difficult to model complex spatial-temporal dependency.
    Temporal dependency includes short-term dependency and long-term dependency, and the latter is often overlooked.
    Spatial dependency can be divided into two parts: distance-based spatial dependency and hidden spatial dependency. 
    To model complex spatial-temporal dependency, we propose a novel framework for traffic forecasting, named Spatial-Temporal Graph Convolutional Gated Recurrent Network (STGCGRN).
    We design an attention module to capture long-term dependency by mining periodic information in traffic data.
    We propose a Double Graph Convolution Gated Recurrent Unit (DGCGRU) to capture spatial dependency, which integrates graph convolutional network and GRU.
    The graph convolution part models distance-based spatial dependency with the distance-based predefined adjacency matrix and hidden spatial dependency with the self-adaptive adjacency matrix, respectively.
    Specially, we employ the multi-head mechanism to capture multiple hidden dependencies.
    In addition, the periodic pattern of each prediction node may be different, which is often ignored, resulting in mutual interference of periodic information among nodes when modeling spatial dependency.
    For this, we explore the architecture of model and improve the performance.
    Experiments on four datasets demonstrate the superior performance of our model.
\end{abstract}

\section{Introduction}
Recently, many cities have been committed to developing intelligent transportation systems (ITS) . 
As an important part of ITS, traffic forecasting has attracted tremendous attention from academia and industry. 
From the perspective of traffic managers, traffic forecasting can help reduce congestion and provide early warning for safety accidents. 
From the traveler's perspective, traffic forecasting can help plan travel routes and improve travel efficiency.

% A lot of studies have investigated traffic forecasting for decades. 
% Some studies apply traditional statistic-based methods, such as
% history average (HA), auto regressive integrated moving average 
% (ARIMA), vector auto regression (VAR), support vector regression 
% (SVR). However, these methods require the features designed 
% by experts rather than mined from the data. Besides, most of these
% methods are linear and do not well capture the complex non-linear 
% spatial-temporal dependency.
Traffic forecasting is a time series forecasting problem, using the past traffic data to predict the data in the future.
Early works simply deployed the classic time series analysis models, e.g., history average (HA), vector 
auto regression (VAR), and auto regressive integrated moving average 
 (ARIMA)\cite{Ahmed1979ANALYSISOF}. These models can only model the linear dependency in the 
data, while the traffic data has complex nonlinear relationship, 
so the performance of these linear models is poor in the real scenario. Therefore, researchers shift to traditional machine learning methods, such as support vector regression (SVR)\cite{Wu2004TraveltimePW} and k-nearest neighbors (KNN)\cite{Davis1991NonparametricRA}. 
However, their performance relies heavily on the handcrafted features, which need to be designed by domain experts.
Recently, deep learning models have become the mainstream choice for traffic forecasting due to their automatic feature extraction ability and better empirical performance.

In deep learning, the temporal dependency of traffic data can be modeled with recurrent neural networks (RNNs)\cite{fu2016using,yu2017deep} or temporal convolution modules\cite{zhang2017deep,Yu2018SpatioTemporalGC,Wu2019GraphWF}.
However, such approaches only mines short-term dependency but ignores long-term dependency.
Traffic data has strong daily and weekly periodicity, which is crucial for traffic forecasting. 
We can mine the periodicity in traffic data to model long-term dependency in traffic data.

The spatial dependency can be captured by convolutional neural network (CNN)\cite{Fang2019GSTNetGS,lin2019deepstn+,yao2019revisiting} or graph convolutional networks (GCN)\cite{Wu2019GraphWF,wang2020traffic,Bai2020AdaptiveGC}. 
CNN can be applied to grid-based maps to capture spatial dependency. 
In this case, the data needs to be divided into $H\times W$ equal size grids, where $H$ and $W$ represent the height and width of the grid-based map respectively. 
However, CNN cannot handle non-Euclidean data, which is a more common form when describing road topology in real scenarios, with the predicted positions as nodes and the roads as edges.
GCN can process non-Euclidean data, calculating the weight of the edge by the distance between the nodes, obtaining a distance-based predefined adjacency matrix, and then feeding it into GCN to capture distance-based spatial dependency\cite{Li2018DiffusionCR,Yu2018SpatioTemporalGC}. 
Besides, the self-adaptive adjacency matrices can be used to model hidden spatial dependency\cite{Wu2019GraphWF,wang2020traffic,Bai2020AdaptiveGC}. 
However, there may be multiple hidden spatial dependencies, which have not been considered in existing works. 
In addition, the periodic patterns of each prediction node may be different, which need to be taken into account when modeling the spatial dependency to prevent the periodic information of different nodes from interfering with each other.

To solve the above problems, we propose Spatial-Temporal Graph Convolutional Gated Recurrent Network (STGCGRN) to improve the accuracy of traffic forecasting. Our contributions are as follows:

\begin{itemize}
    \item We propose a new framework to better exploit temporal and spatial dependency: We capture long-term temporal dependency by explicitly introducing daily and weekly periodic data, which are fed into the attention module with a sliding window. The DGCGRU layer is designed to extract the distance-based spatial dependency and multiple hidden spatial dependencies, which are captured by the predefined distance-based adjacency matrix and the multi-head self-adaptive adjacency matrix, respectively.
    \item We explore the model structure and find that mining the periodic information of a single node, and then performing spatial dependency modeling can reduce the interference of periodic information among different nodes.
    \item We conduct experiments on four datasets. The results show that our model outperforms the baseline methods. We also conduct ablation experiments to evaluate the impact of each component of our model on the performance.
\end{itemize}
\section{Related Work} \label{relatedwork}
\subsection{Graph Convolutional Neural Network}
Graph convolutional neural network is a special form of CNN generalized to non-Euclidean data, which can be applied to different tasks and domains, such as node classification \cite{Kipf2017SemiSupervisedCW}, graph classification \cite{ying2018hierarchical}, link prediction \cite{zhang2018link}, node clustering \cite{wang2017mgae}, spatial-temporal graph forecasting, etc.
Bruna et al.\cite{Bruna2014SpectralNA} is the first work to generalize the convolution operation to non-Euclidean data, and proposes a spatial method and a spectral method.

Spatial-based approaches aggregate neighborhood feature informations in the spatial domain to extract a node’s high-level representation.
DCNNs\cite{Atwood2016DiffusionConvolutionalNN} extend convolutional neural networks (CNNs) to general graph-structured data by introducing a ‘diffusion-convolution’ operation.
Message Passing Neural Network (MPNN)\cite{gilmer2017neural} describes a general framework for supervised learning on graphs, applying graph convolution to supervised learning of molecules.
GraphSage\cite{Hamilton2017InductiveRL} proposes a general inductive framework to efficiently generate node embeddings for previously unseen data, sampling and aggregating features from the node’s local neighborhood.

Spectral-based approaches implement the convolution operation in the spectral domain with graph Fourier transforms.
ChebyNet\cite{Defferrard2016ConvolutionalNN} achieves fast localized spectral filtering by introducing C order Chebyshev polynomials parametrization.
GCN\cite{Kipf2017SemiSupervisedCW} further reduces computational complexity of ChebNet by limiting C=2.
\subsection{Traffic Forecasting}
As an important component of ITS, traffic forecasting has received wide attention for decades.
Most of the early traffic forecasting works are based on some classic time series analysis models. ARIMA \cite{Ahmed1979ANALYSISOF} is used to predict the short-term freeway traffic data.
Subsequently, many variants of ARIMA are applied to this task, such as KARIMA \cite{van1996combining}, subset ARIMA \cite{Lee1999ApplicationOS}, ARIMAX \cite{Williams2001MultivariateVT}, etc.
Traffic data contains complex spatial-temporal dependency. However, the aforementioned works only consider the linear relationship in the temporal dimension, which is obviously insufficient. 
Researchers apply traditional machine learning algorithms to traffic forecasting, including SVR \cite{Wu2004TraveltimePW}, KNN \cite{Davis1991NonparametricRA}, and so on.
These methods need handcrafted features, which require expert experience in related fields

Due to the advantages of deep learning in solving nonlinear problems and automatically extracting features, scholars pay more attention to deep learning.
RNN \cite{fu2016using,yu2017deep} and CNN \cite{zhang2017deep,Yu2018SpatioTemporalGC,Wu2019GraphWF} are adopted to capture temporal dependency. For example, Graph Wavenet \cite{Wu2019GraphWF} uses dilated casual convolutions to model the temporal dependency to increase the receptive field. 
To model the spatial dependency, Some works \cite{Fang2019GSTNetGS,lin2019deepstn+,yao2019revisiting} divide the traffic data into grid-based map, and extract spatial information by CNN.
However, due to natural limitations, CNN can not handle non-Euclidean data. For this reason, researchers apply GCN to the field of traffic forecasting.
DCRNN \cite{Li2018DiffusionCR} captures the spatial dependency using bidirectional random walks on the graph.
STGCN \cite{Yu2018SpatioTemporalGC} employs a generalization of Chebnet to capture the spatial dependency of traffic data.
These methods only utilize the predefined graph structures, which is not sufficient.
Some works \cite{Wu2019GraphWF,wang2020traffic,Bai2020AdaptiveGC} calculate similarities among learnable node embeddings to capture the hidden spatial dependency in the data.
DGCRN \cite{li2021dynamic} filters the node embeddings and then uses them to generate dynamic graph at each time step.
% ASTGNN \cite{guo2021learning} employs self-attention to capture the spatial correlations in a dynamic manner.
However, these works do not adequately model spatial dependency and ignore periodic information.

\section{Preliminaries} \label{preliminaries}
\begin{itemize}
\item Definition 1: Traffic Network. The traffic network is an undirected graph $G= (V,E,A)$, where $V$ is a set consists of $N$ nodes, $E$ is a set of edges and $A \in \mathbb{R}^{N \times N}$ is the adjacency matrix representing the nodes distance-based proximity.
\item Definition 2: Traffic Signal Matrix. Traffic signal matrix can be denoted as a tensor $X_{t}\in \mathbb{R}^{N\times C}$, where $C$ is the number of traffic features of each node (e.g., the speed, volume), $t$ denotes the time step.
\end{itemize}
The problem of traffic forecasting can be described as: given the traffic network $G= (V,E,A)$ and $P$ steps traffic signal matrix $X_{ (t-P):t}= (X_{t-P}, X_{t-P+1}, ..., X_{t-1})\in \mathbb{R}^{P \times N \times C}$, learning a function $f$ which maps $X_{ (t-P):t}$ to next $Q$ steps traffic signal matrix $X_{t: (t+Q)}= (X_{t}, X_{t+1}, ..., X_{t+Q-1})\in \mathbb{R}^{P \times N \times C}$, represented as follows:
\begin{equation}\label{eq_1}
    X_{t: (t+Q)} = f (X_{ (t-P):t}; G)
\end{equation}

\section{Spatial-Temporal Graph Convolutional Gated Recurrent Network} \label{method}

\begin{figure}[t]
    %\label{ablation_res}
    \centering
    \subfloat[Model Framework]{%
        \includegraphics[width=1\columnwidth]{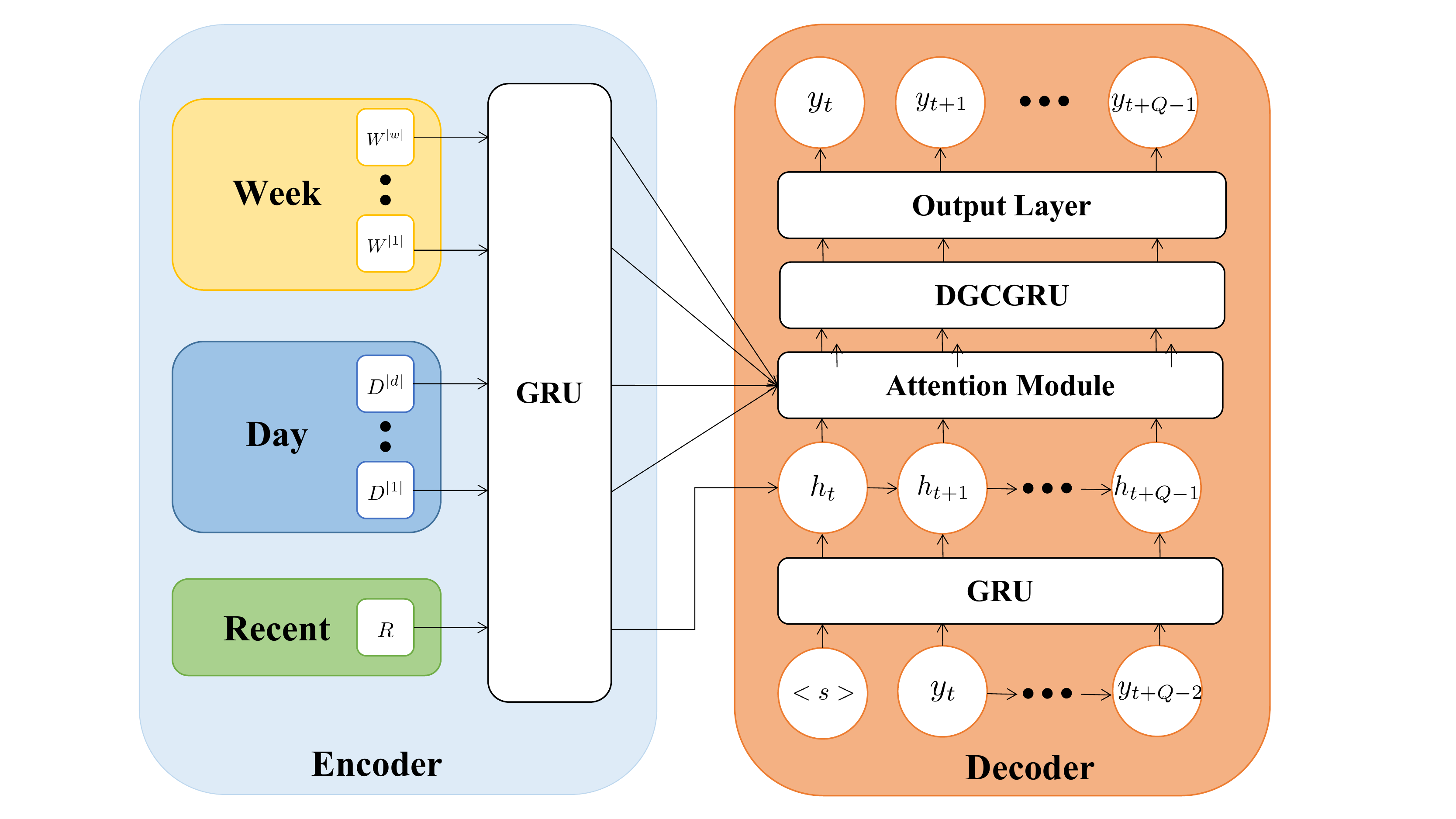}
    }\\
    \subfloat[Encode Traffic Data]{%
    \includegraphics[trim=50 200 50 30,width=1\columnwidth]{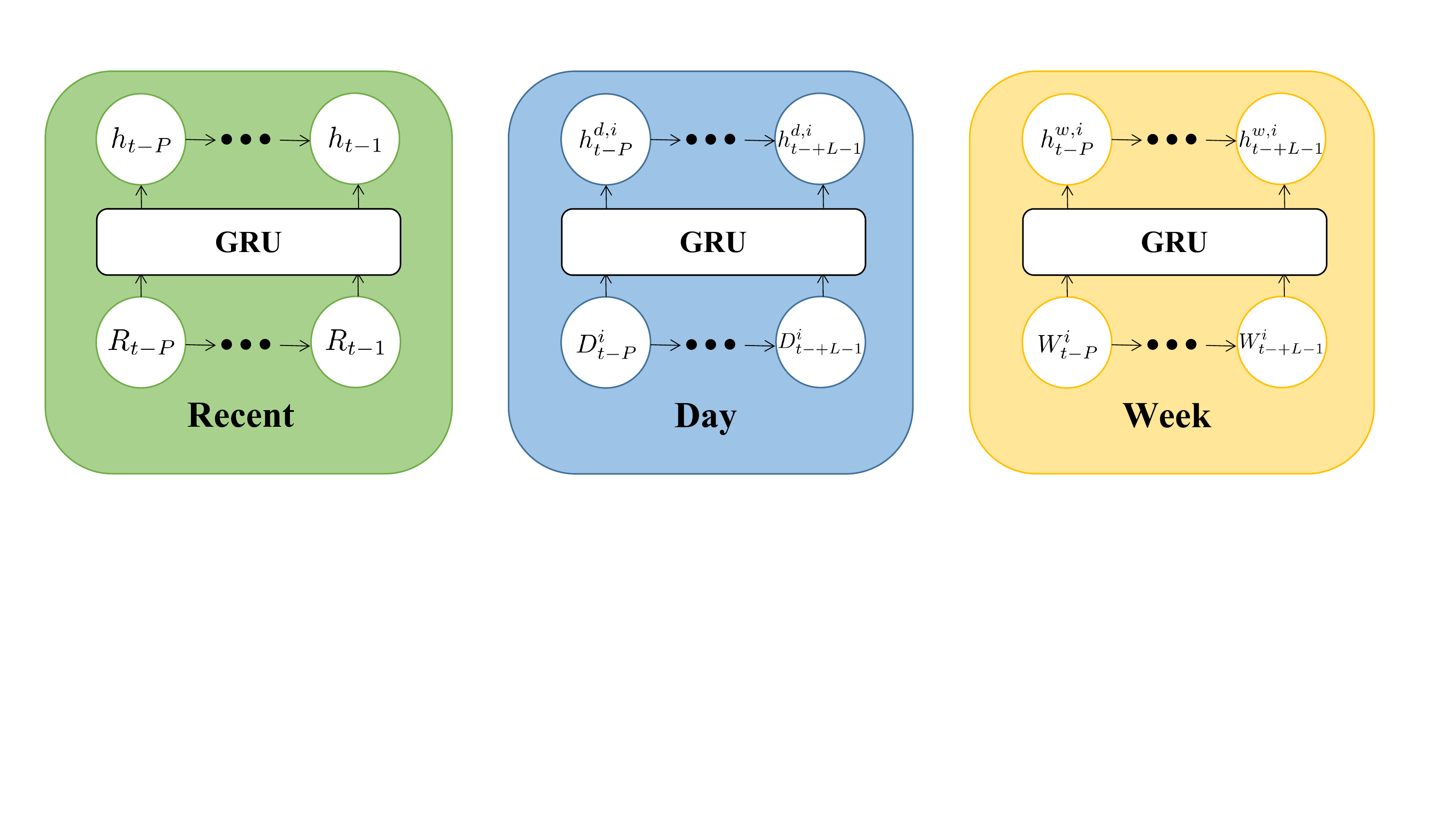}
    } 
    \caption{ (a) is the architecture of the STGCGRN. It contains an encoder and a decoder. The former simply uses GRU, and the decoder consists of a GRU layer, an attention module, a DGCGRU layer, and an output layer.
    The GRU layer generates hidden state vectors at each time step, modeling the short-term temporal dependency.
    The attention module mines periodic information in the traffic data to capture long-term dependence. 
    The DGCGRU layer models spatial dependency. 
    The purpose of placing the DGCGRU layer behind the attention module is to prevent the periodic information of nodes from interfering with each other. Subsequent experiments have also proved the effectiveness of this point.
    The output layer is a fully-connected layer.~ (b) is the details of the encoder.}
    \label{model}
\end{figure}

% \begin{figure}[t]
%     %\label{ablation_res}
%     \centering
%     \begin{minipage}[c]{1\columnwidth}
% 		\centering
% 		\includegraphics[width=\columnwidth]{pics/model1.pdf}
% 		\centerline{(a)Model Framework}
% 		%\label{fig_E2_1}
% 	\end{minipage} \\

%     \begin{minipage}[c]{1\columnwidth}
% 		\centering
% 		\includegraphics[trim=50 200 50 0,width=1\columnwidth]{pics/model1_1.pdf}
% 		\centerline{(b)Encode Traffic Data}
% 		%\label{fig_E2_1}
% 	\end{minipage}
%     \caption{ (a) is the architecture of the STGCGRN. It contains an encoder and a decoder. The former simply uses GRU, and the decoder consists of a GRU, an attention module, a DGCGRU layer, and a output layer.
%     The GRU layer generates hidden state vectors at each time step, modeling the short-term temporal dependency.
%     The attention module mines periodic information in the traffic data to capture long-term dependence. 
%     And the DGCGRU layer models spatial dependency. 
%     The purpose of placing the DGCGRU layer behind the attention module is to prevent the periodic information between nodes from interfering with each other. Subsequent experiments have also proved the effectiveness of this point.
%     The output layer is a fully-connected layer.~ (b) is the is the details of the encoder}
%     \label{model}
% \end{figure}

We design the STGCGRN based on the sequence-to-sequence architecture to capture spatial-temporal dependency.
The architecture of our model is shown in Figure~\ref{model}.

\subsection{Periodic~Data~Construct} \label{periodi_data_construct}
Before introducing the model, we first explain the input data structure.

In order to model periodicity, we replace the traffic signal matrix $X_{ (t-P):t}$ in Equation~\ref{eq_1} with three time granularity inputs: recent data, daily periodicity, and weekly periodicity, denoted by $R, D, W$ respectively.
Details about the three parts are as follows:

 (1) The recent traffic data:

The recent traffic data is a segment of traffic data before the forecast period.
Intuitively, the recent traffic data have a great impact on the traffic data in the future.
\begin{equation}\label{eq_2}
\begin{split}
    &R = (R_{t-P}, R_{t-P+1}, ..., R_{t-1})\in \mathbb{R}^{P \times N \times C}
    \\
    &R_{t} = (r_{1,t}, r_{2,t}, ..., r_{N,t})\in \mathbb{R}^{N \times C}
\end{split}
\end{equation}
where $r_{i,t}$ means the traffic data of node $i$ at time step $t$.

 (2) The daily periodic traffic data:

The daily periodic part is the data of the same period of the previous |d| days. 
Since traffic data is determined by people's daily routine, there are often obvious fixed patterns, such as morning and evening peaks.
The daily periodic traffic data is introduced for the pattern.
\begin{equation}\label{eq_3}
    \begin{split}
        &D = (D^{|d|},D^{|d|-1}, ..., D^{1})\in \mathbb{R}^{|d| \times (P+L) \times N \times C}
        \\
        &D^{i} = (D_{t-P}^{i}, D_{t-P+1}^{i}, ..., D_{t+L-1}^{i})\in \mathbb{R}^{ (P+L) \times N \times C}
        \\
        &D_{t}^{i} = (d_{1,t-i*l_{d}}, d_{2,t-i*l_{d}}, ..., d_{N,t-i*l_{d}})\in \mathbb{R}^{N \times C}
    \end{split}
\end{equation}
where $l_{d}$ means the number of samples per day, $|d|$ represents the number of days used, $L$ will be introduced later.

 (3) The weekly periodic traffic data:

The weekly periodic part is the data of the same period of the previous |w| weeks, which is to model the weekly periodicity.

\begin{equation}\label{eq_4}
    \begin{split}
        &W = (W^{|w|},W^{|w|-1}, ..., W^{1})\in \mathbb{R}^{|w| \times (P+L) \times N \times C}
        \\
        &W^{i} = (W_{t-P}^{i}, W_{t-P+1}^{i}, ..., W_{t+L-1}^{i})\in \mathbb{R}^{ (P+L) \times N \times C}
        \\
        &W_{t}^{i} = (w_{1,t-i*l_{w}}, w_{2,t-i*l_{w}}, ..., w_{N,t-i*l_{w}})\in \mathbb{R}^{N \times C}
    \end{split}
\end{equation}
where $l_{w}$ means the number of samples per week, $|w|$ represents the number of weeks used.

After introducing the period information, we can update the Equation~\ref{eq_1} to the following form:

\begin{equation}\label{eq_5}
    X_{t: (t+Q)} = f (R,D,W; G)
\end{equation}

\subsection{Attention~Module} \label{attention_module}
% We use Gated Recurrent Unit (GRU) as the basic unit for processing traffic data, which alleviates the vanishing gradient problem in traditional Recurrent Neural Network (RNN) and is less computationally expensive than Long Short-Term Memory (LSTM) network.

% We send the recent traffic data $R$, the daily periodic traffic data $D$, the weekly periodic traffic data $W$ into the GRU, the formula is as follows:
% \begin{equation}\label{eq_6}
%     h_{i,t} = GRU (x_{i,t}, h_{i,t-1}) \in \mathbb{R}^{d_{h}}
% \end{equation}
% where $h_{i,t}$ means the hidden state vector of node $i$ at time step $t$, and $d_{h}$ represents the dimension of $h_{i,t}$.
% In addition, $R$ uses a GRU cell alone, and $D$ and $W$ share a GRU cell.

\begin{figure}[t]
    \centering
    \includegraphics[trim=50 0 130 0,width=1\columnwidth]{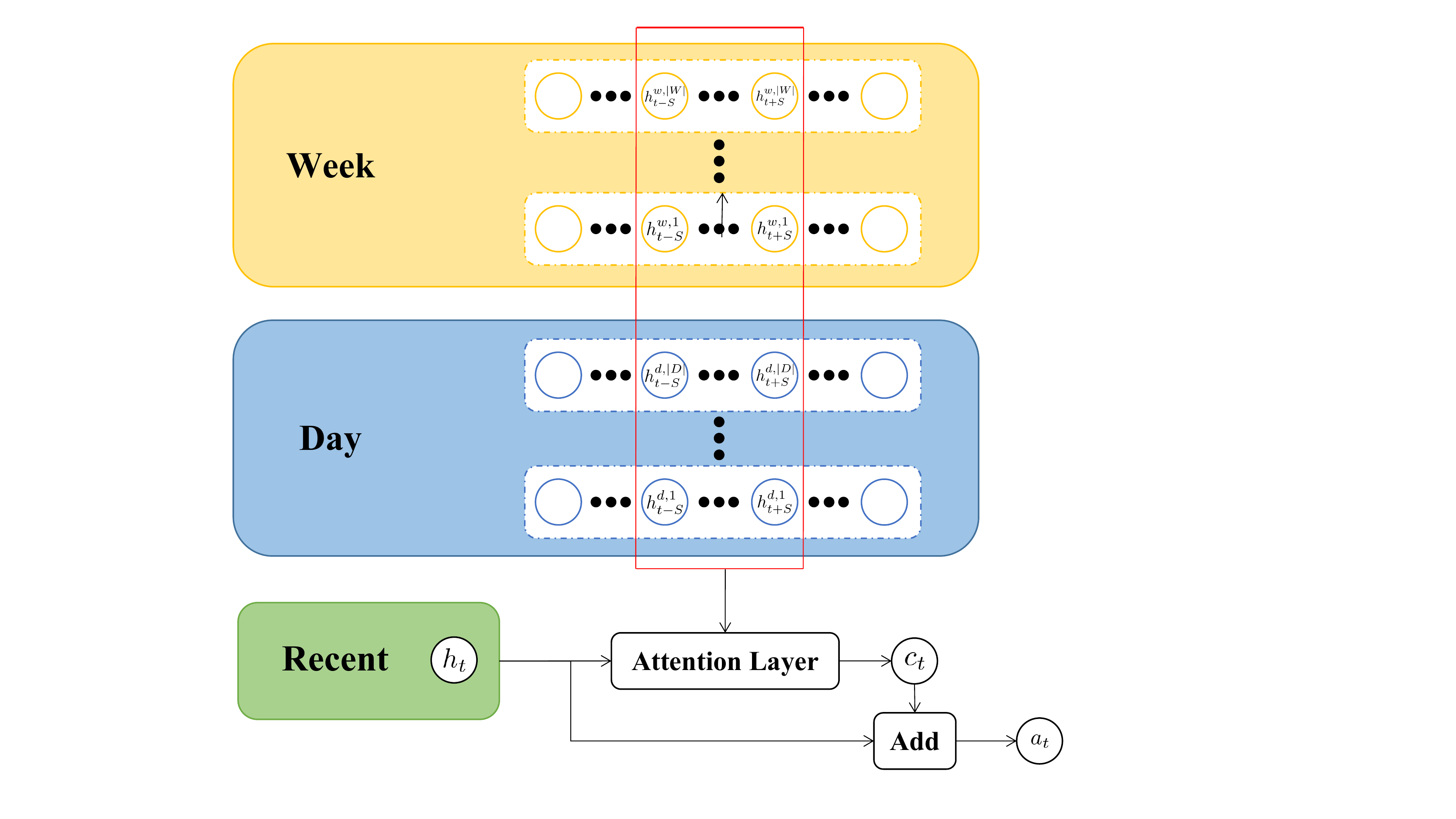}
    \caption{Attention Module}
    \label{attention_module_fig}
\end{figure}

Weekly periodicity can be obtained through $W$. For example, the traffic data from last Saturday is helpful for the traffic forecast of this Saturday.
However, due to the temporal shifting of periodicity\cite{Yao2019RevisitingSS}, daily periodicity cannot depend only on the same moment of the previous days.
For example, yesterday's traffic peak occurred at 5:00 pm, but today's traffic peak may occur at 5:10 pm. So it is incorrect to use the traffic data of 5:10 pm yesterday as daily periodic data.
The traffic data at 5:00 pm yesterday is a better choice.
That is to say, the daily periodicity is not strictly periodic, and there will be a certain time deviation.
But in general, this deviation is not very large.
To cope with the deviation, we consider several time steps adjacent to the predicted time step.
We open a time window around the prediction moment, and only focus on the moment within the window.
The size of the window is $2*S+1$.
We introduce the variable $L$ when we construct $D$ and $W$. 
According to the above analysis, we let $L=Q+S$.

The attention module is shown in Figure~\ref{attention_module_fig}.
We apply attention mechanism to the data within the window to capture daily periodicity.
The decoder generates a hidden state vector $h_{i,t}$ for node $i$ at time step $t$.
Meanwhile, we generate a time window centered on $t$ for the hidden state vectors of the daily periodic and weekly periodic data, and then concatenate the hidden state vectors in the window to get $h_{i,t}^{period}$:
\begin{equation}\label{eq_7}
    \begin{split}
        &h_{i,t}^{period} = (h^{d,|d|}_{i,t,S}, ..., h^{d,1}_{t,S}, h^{w,|w|}_{i,t,S}, ..., h^{w,1}_{t,S})
        \\
        &\in \mathbb{R}^{ ( (|w|+|d|)\times (2*S+1))\times d_{h}}
        \\
        &h^{d,j}_{i,t,S} = (h_{i,t-S-j*l_{d}}, ..., h_{i,t+S-j*l_{d}})\in \mathbb{R}^{ (2*S+1) \times d_{h}}
        \\
        &h^{w,j}_{i,t,S} = (h_{i,t-S-j*l_{w}}, ..., h_{i,t+S-j*l_{w}})\in \mathbb{R}^{ (2*S+1) \times d_{h}}
    \end{split}
\end{equation}
where $d_{h}$ is the dimension of the hidden vector.

In order to simplify the representation, we use $h_{p}$ to represent the hidden state vector in $h_{i,t}^{period}$.
We use the attention mechanism to assign a weight $w_{p}$ to each $h_{p}$. Then the content vector $c_{i,t}$ is obtained by weighted sum of $h_{p}$, which is defined as:
\begin{equation}\label{eq_8}
    c_{i,t} = \sum_{s} w_{p}h_{p}
\end{equation}
where $w_{p}$ measures the importance of $h_{p}$, which is calculated as follows:
\begin{equation}\label{eq_9}
    w_{p} = \frac{exp (score (h_{i,t},h_{p}))}{\sum_{s}exp (score (h_{i,t},h_{p}))}
\end{equation}
where $h_{i,t}$ means the hidden state of node $i$ at time step $t$ generated by the GRU layer.

The implementation of the score function refers to \cite{Luong2015EffectiveAT}:
\begin{equation}\label{eq_10}
    score (h_{i,t},h_{p}) = v^{T}tanh (W_{1}h_{i,t}+W_{2}h_{p}+b)
\end{equation}
where $v^{T},W_{1},W_{2},b$ are learnable parameters.

The output of the attention module is shown in the following equation:
\begin{equation}\label{eq_11}
    a_{i,t} = c_{i,t} + h_{i,t}
\end{equation}

To sum up, the weekly periodicity is introduced through the weekly periodic traffic data $W$. 
Then the attention mechanism is applied to the daily periodic data $D$ and weekly periodic data $W$ to capture the daily periodicity.
By processing $R$ with GRU, we can capture short-term dependency in traffic data.
By mining periodicity in traffic data, we can model long-term dependency.

\subsection{DGCGRU}

\begin{figure}[t]
    \centering
    \subfloat[DGCGRU]{%
    \includegraphics[trim=50 0 50 0,width=0.8\columnwidth]{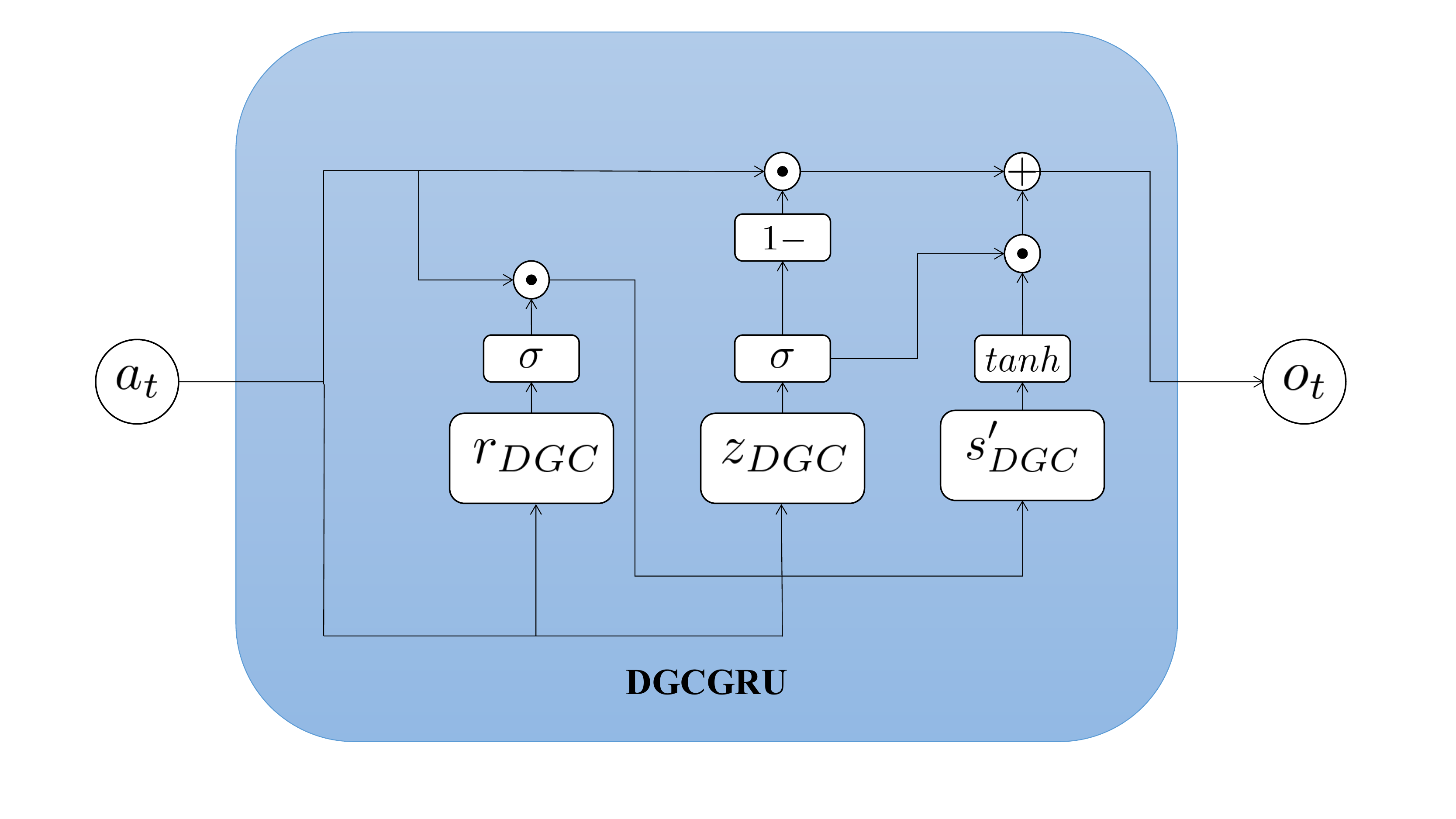}
        %\includegraphics[width=0.45\columnwidth]{pics/model4.jpg}
        %\label{GCGRU Cell}
    }\\
    \subfloat[Double Graph Convolution]{%
    \includegraphics[trim=80 40 80 40,width=0.9\columnwidth]{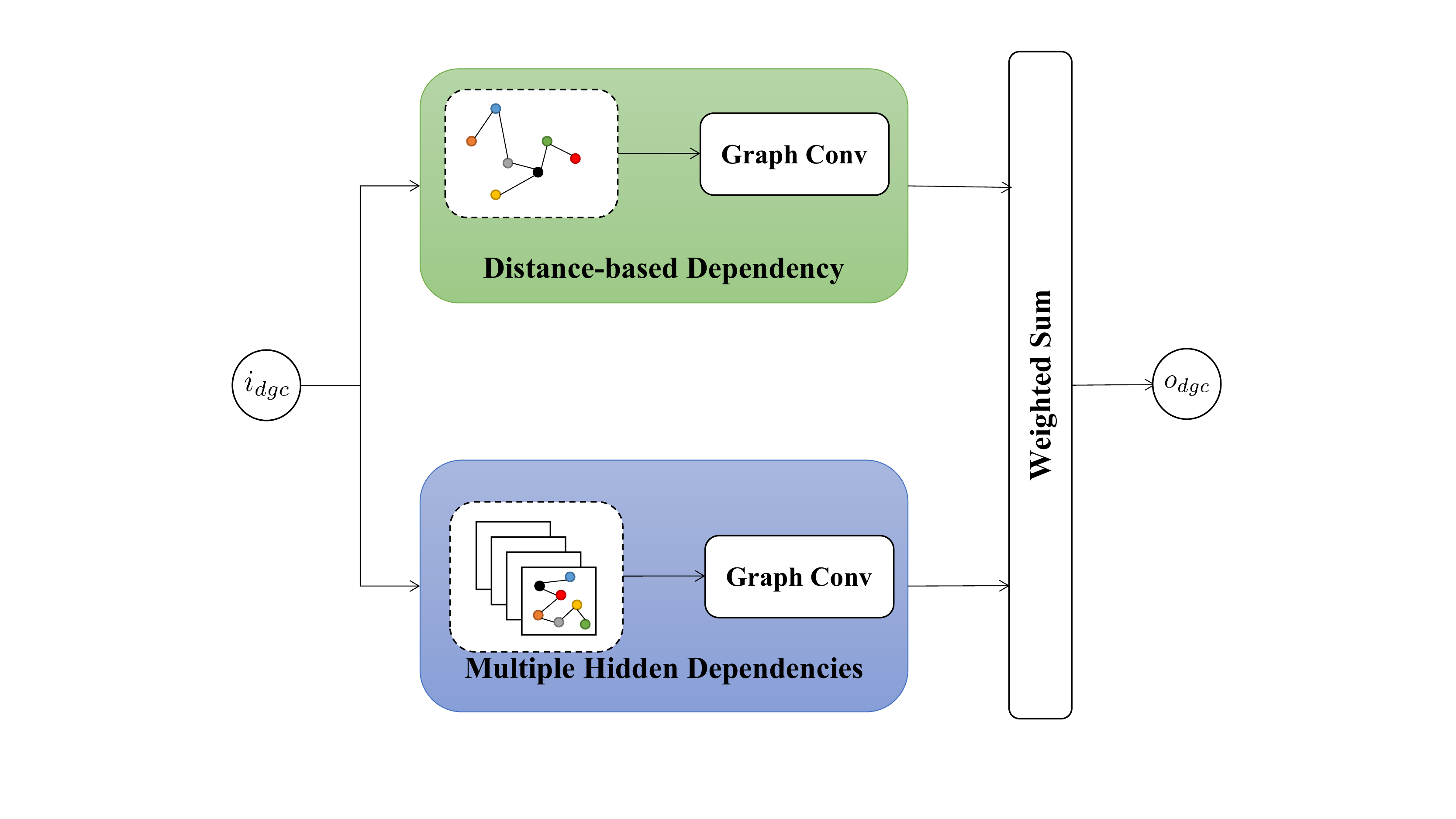}
    %\includegraphics[width=0.45\columnwidth]{pics/model3.jpg}
    %\label{Dual Graph Convolution}
    } 
    \caption{ (a) is the architecture of the Double Graph Convolutional Gated Recurrent Unit (DGCGRU), replacing the full connection layer of gate unit in GRU with double graph convolution operation. (b) is the detail of double graph convolution operation}
    \label{gcgru}
\end{figure}

% \begin{figure}[t]
%     \label{spatial_dependency_module}
%     \centering
%     \includegraphics[width=0.4\columnwidth]{pics/model1.pdf}
%     % \includegraphics[trim=100 0 100 0,width=1\columnwidth]{pics/model3.pdf}
%     \caption{Diagram of {\myfont Multi-Head\,Co-Training} with three heads.
%     }
%     \label{MM}
% \end{figure}

% The decoder generates hidden states $H_{t}= (h_{1,t}, h_{2,t}, ..., h_{N,t}) \in \mathbb{R}^{N \times d_{h}}$ at time step $t$, which contain short-term dependency.
% The attention module generates tensor $S_{t}= (s_{1,t}, s_{2,t}, ..., s_{N,t}) \in \mathbb{R}^{N \times d_{h}}$ by mining periodic information in $H_{t}$, capturing long-term dependency.
The attention module generates tensor $A_{t}= (a_{1,t}, a_{2,t}, ..., a_{N,t}) \in \mathbb{R}^{N \times d_{h}}$. Next, we send $A_{t}$ to the DGCGRU layer to model the spatial dependency from two aspects: the distance-based spatial dependency and the hidden spatial dependency, as shown in Figure~\ref{gcgru}.

\subsubsection{Distance-based~Spatial~Dependency}
Graph Convolution Networks (GCNs) generalize convolution operation to non-Euclidean data.
The idea of GCNs is to learn the representation of nodes by exchanging information between them.
Specifically, given a node, GCNs first generate an intermediate representation of the node by aggregating the representations of its neighbor nodes, and then perform a transformation on the intermediate representation to obtain the node's representation.

Traffic network is usually a graph structure. 
The vector $a_{i,t}$ generated at each time step can be regarded as the signal of the graph.
In general, the closer two predicted points are, the higher the correlation between them.
To take full advantage of the topology of the traffic network, we perform a graph convolution operation on $A_{t}$ based on distance information of the graph at each time step.
The k-hop graph convolution operation is as follows:
\begin{equation}\label{eq_12}
    \begin{split}
        &o^{pre} = \sum_{i=0}^{K}S^{k}W^{k}
        \\
        &S^{k} = S^{k-1}\widetilde{A}^{pre}
        \\
        &S^{0} = i_{dgc}
        \\
        &\widetilde{A}^{pre} = (\widetilde{D}^{pre})^{-1}A^{pre}
        \\
        &\widetilde{D}^{pre}_{i,i} = \sum_{j}A^{pre}_{i,j}
    \end{split}
\end{equation}
where $A^{pre}\in \mathbb{R}^{N \times N}$ is predefined adjacency matrix that contains the distance information among nodes.
$i_{dgc}$ is the input of the double graph convolution operation.
$W^{k}\in \mathbb{R}^{d_{h} \times d_{h}}$ are learnable parameters. 
$K$ is the number of hops for graph convolution.
$o^{pre}$ is the output of this submodule.

\subsubsection{Hidden~Spatial~Dependency}
The distance-based predefined adjacency matrix lacks hidden spatial dependency that are usually directly related to prediction tasks.

To solve this problem, some works \cite{Bai2020AdaptiveGC,Wu2019GraphWF} adopt a self-adaptive adjacency matrix to automatically infer the hidden spatial dependency between nodes, which is learned through stochastic gradient descent.
However, these works ignore that there may be multiple hidden spatial dependencies among nodes.

In our work, we combine the self-adaptive adjacency matrix with multi-head mechanism to model multiple hidden dependencies.
We randomly initialize two sets of node embeddings $E_{1},E_{2}\in \mathbb{R}^{N \times n_{head} \times d_{e}}$, where $n_{head}$ is the number of heads, and $d_{e}$ represents the dimension of the embedding of each head.
Next, we calculate the hidden dependencies among nodes according to the following formula:
\begin{equation}\label{eq_13}
    \widetilde{A}^{adp}_{i} = softmax (\frac{ReLU (E_{1,i}E_{2,i}^{T})}{d_{e}})
\end{equation}
where $i$ indicates the $i$th head, and the ReLU function is used to eliminate some weak connections.
Through the multi-head mechanism, we divide the embedding into multiple subspaces, and learn the corresponding self-adaptive graph in each subspace.
It is worth noting that instead of generating the adjacency matrix $A^{adp}$ and then calculating the Laplacian matrix, we directly obtain the normalized self-adaptive adjacency matrix $\widetilde{A}^{adp}_{i}$ through the softmax function, avoiding unnecessary calculations.

We replace $A^{pre}$ in Equation~\ref{eq_12} with $\widetilde{A}^{adp}_{i}$, and then perform a graph convolution operation to obtain $o^{adp}_{i}$.
The output $o^{adp}$ models multiple hidden spatial dependency by averaging the $o^{adp}_{i}$ of each head.

\subsubsection{Multi-Component~Fusion}
The final output of the spatial dependency module is as follows:
\begin{equation}\label{eq_15}
    o_{dgc} = w^{pre}o^{pre}+w^{adp}o^{adp}
\end{equation}
where $w^{pre},w^{adp}$ are hyper parameters
% \begin{figure}[t]
%     \centering
%     \includegraphics[width=1\columnwidth]{pics/model1.pdf}
%     \caption{Diagram of {\myfont Multi-Head\,Co-Training} with three heads.
%     Images are fed into a shared module (blue box) followed by three classification heads (green boxes). 
%     Among them, weakly augmented images (orange lines) are for pseudo-labeling.
%     The pseudo-labels are to guide the predictions on strongly augmented examples (red line).
%     Here, pseudo-labels for the bottom head are generated and selected according to the other two heads’ predicted classes on the weakly augmentation images.
%     \textit{Note that only the co-training process of the bottom head is shown here.}
%     The weakly and strongly augmented images are in fact simultaneously fed into all three heads.
%     }
%     \label{MM}
% \end{figure}

\section{Experiments} \label{experiments}
\subsection{Datasets}
To evaluate the performance of our model, we conduct experiments on four datasets.

\subsubsection{Dataset~Description}
We use the following four datasets: PEMS03, PEMS04, PEMS07, and PEMS08, which are constructed from four districts, respectively in California.
All data is collected from the Caltrans Performance Measurement System(PeMS)\cite{chen2001freeway}.
The traffic data is aggregated into every 5-minute interval, which means every sensor contains 12 traffic data points per hour.
The detailed information of the dataset is shown in Table \ref{pems}.
\begin{table}
    \centering
    \begin{tabular}{lccc} \toprule
        Datasets    & \#Nodes    & \#Edges       &  Time range\\ \midrule % \makecell{Multi-Head\\Co-Training} 
        PEMS03      & 358        & 547           & 9/1/2018-11/30/2018\\
        PEMS04      & 307        & 340           & 1/1/2018-2/28/2018\\
        PEMS07      & 883        & 866           & 5/1/2017-8/31/2017 \\
        PEMS08      & 170        & 295           & 7/1/2016-8/31/2016 \\
        \bottomrule
    \end{tabular}
    \caption{Dataset description.}
    \label{pems}
    %\label{params}
\end{table}
\subsubsection{Data~Preprocessing}
We split all datasets into training set, validation set and test set with the ratio of 6:2:2.
% We normalize the data with:
% \begin{equation}\label{eq_16}
%     X_{norm} = \frac{X-\mu}{\sigma}
% \end{equation}
% where $\mu$ is the mean of the data, and $\sigma$ is the standard deviation.

We compute the distance-based predefined adjacency matrix $A^{pre}$ using thresholded Gaussian kernel \cite{Shuman2013TheEF}:
\begin{equation}\label{eq_17}
    \begin{aligned}
    w_{ij}=\begin{cases}
        exp (-\frac{dist (i,j)^{2}}{\sigma^{2}}) &\text{ if } dist (i,j)^{2}\leq \kappa \\
        0 &\text{otherwise }  
       \end{cases}
    \end{aligned}
\end{equation}
where $w_{ij}$ is the weight of the edge between node $i$ and node $j$, $dist (i,j)$ is distance  between node $i$ and node $j$.
$\sigma$ is the standard deviation of distances and $\kappa$ is the threshold.

\subsection{Baseline~Methods}
%We compare STGCGRN with the following models:
\begin{itemize}
    \item MLP: Multilayer Perceptron uses two fully connected layers for traffic forecasting.
    \item CNN: Convolutional Neural Network performs convolution operations in the temporal dimension to capture temporal correlations.
    \item GRU: Gate Recurrent Unit with fully connected layer is powerful in modeling time series. 
    \item DCRNN\cite{Li2018DiffusionCR}: Diffusion Convolutional Recurrent Neural Network captures the spatial dependency using bidirectional random walks on the graph, and the temporal dependency using the encoder-decoder architecture.
    \item STGCN\cite{Yu2018SpatioTemporalGC}: Spatio-Temporal Graph Convolutional Networks combines graph convolutional layers and convolutional sequence learning layers, to model spatial and temporal dependency.
    \item ASTGCN (r)\cite{Guo2019AttentionBS}: Attention based Spatial-Temporal Graph Convolution Network (r) adopts the spatial-temporal attention mechanism to effectively capture the dynamic spatial-temporal correlations in traffic data. Only recent components is used.
    \item Graph WaveNet\cite{Wu2019GraphWF}: Graph WaveNet capture the hidden spatial dependency by developing a self-adaptive dependency matrix, and enlarges the receptive field a stacked dilated 1D convolution component.
    \item AGCRN\cite{Bai2020AdaptiveGC}: Adaptive Graph Convolutional Recurrent Network employs adaptive graph to capture node-specific spatial and temporal correlations.
    \item STSGCN\cite{song2020spatial}: Spatial-Temporal Synchronous Graph Convolutional Network captures the complex localized spatial-temporal correlations through an elaborately designed spatial-temporal synchronous modeling mechanism.
    \item DGCRN\cite{Li2021DynamicGC}: Dynamic Graph Convolutional Recurrent Network employs a generation method to model fine topology of dynamic graph at each time step.
    \item ASTGNN\cite{guo2021learning}: Attention based Spatial-Temporal Graph Neural Network employs self-attention to capture the spatial correlations in a dynamic manner while also introducing periodic information.
\end{itemize}

\subsection{Experiment~Settings}
\subsubsection{Metrics}
We deploy Mean Absolute Error (MAE), Mean Absolute Percentage Error (MAPE), Root Mean Square Error (RMSE) as evaluation metrics to measure the performance of the models.

\subsubsection{Optimizer~\&~Early Stopping}
We implement the STGCGRN model using pytorch.
We choose MAE as the loss function and Adam as the optimizer.
We employ early stopping to avoid overfitting. 
The model with the best loss of the validation set is selected as the evaluation model.
We repeat the experiment 5 times and report the average value of evaluation metrics. 
\subsubsection{Hyper parameters}
Hyper parameters are tuned on the validation set.
One hour historical data is used to construct $R$. 
For $D, W$, we let $|d|=1, |w|=1$, and $S=3$.
For multi-component fusion, we let $w^{pre}=0.1,w^{adp}=0.9$.
We set the number of heads of the node embedding that generates the self-adaptive adjacency matrix to 8.
%We predict the traffic data over one hour in the future, that is to say, the predicted sequence length is 12.
The learning rate is set to 0.001, and the batch size is set to 16.
\subsection{Experiment~Results} \label{results}
We compare our model with the baseline methods on four datasets. Table \ref{all_results} shows the results of traffic forecasting performance over the next hour.
% Our model achieves the best performance on three other datasets.% except PEMS03.
%In the PEMS03, the MAPE and RMSE of STGCGRN are optimal, except that the RMSE is slightly worse than that of MTGNN.

GRU outperforms MLP and CNN, which means that RNNs have a natural advantage in dealing with sequence data.
MLP, CNN, and GRU only take short-term temporal dependency into consideration, ignoring spatial dependency in traffic data. 
In contrast, other models, such as DCRNN, AGCRN, DGCRN, etc., exploit spatial dependency from different perspectives, performing better than MLP, CNN, and GRU.
DCRNN, STGCN, ASTGCN capture distance-based spatial dependency through the predefined adjacency matrix, and achieve a good performance.
AGCRN benefits a lot from the self-adaptive adjacency matrix, mining hidden spatial dependency.
Graph Wavenet simultaneously model distance-based spatial dependency and hidden spatial dependency.
% However, the above models fail to model the dynamic characteristics of the traffic data, which limits the performance of the models.
By modeling dynamic spatial dependencies, DGCRN further improves model performance.
By introducing periodic information, ASTGNN outperforms other baselines.
Our model avoids the interference of periodic information among nodes while modeling spatial-temporal dependence, further enhancing the performance.

\begin{table*}[t]
    \centering
    % \resizebox{\textwidth}{!}{%
    \resizebox{\linewidth}{!}{
    \begin{tabular}{lcccccccccccccc}
    \toprule
    Datasets     &Metric                          &MLP                                                                         &CNN                                                                         &GRU                                                                         &DCRNN                                                                       &STGCN                                                                       &ASTGCN (r)                                                                   &Graph Wavenet                                                               &AGCRN                                                                       &STSGCN                                                                      &DGCRN                                                                       &ASTGNN                                                                               &STGCGRN      \\ \midrule                                                                                                              %  
    PEMS03       &\thead{MAE \\ MAPE (\%) \\RMSE}  &\thead{\RS{20.50$\pm$0.12} \\ \RS{24.43$\pm$3.21} \\\RS{31.67$\pm$0.08}}    &\thead{\RS{20.99$\pm$0.15} \\ \RS{23.05$\pm$2.72} \\\RS{33.03$\pm$0.21}}    &\thead{\RS{20.56$\pm$0.18} \\ \RS{22.84$\pm$2.20} \\\RS{33.09$\pm$0.14}}    &\thead{\RS{18.39$\pm$0.39} \\ \RS{20.22$\pm$2.83} \\\RS{30.56$\pm$0.17}}    &\thead{\RS{18.28$\pm$0.39} \\ \RS{17.52$\pm$0.32} \\\RS{30.73$\pm$0.78}}    &\thead{\RS{17.85$\pm$0.45} \\ \RS{17.65$\pm$0.79} \\\RS{29.88$\pm$0.65}}    &\thead{\RS{14.79$\pm$0.08} \\ \RS{14.32$\pm$0.24} \\\RS{25.51$\pm$0.17}}    &\thead{\RS{15.58$\pm$0.03} \\ \RS{15.19$\pm$0.36} \\\RS{27.50$\pm$0.23}}    &\thead{\RS{17.48$\pm$0.15} \\ \RS{16.78$\pm$0.20} \\\RS{29.21$\pm$0.56}}    &\thead{\RS{14.96$\pm$0.07} \\ \RS{15.35$\pm$0.32} \\\RS{26.44$\pm$0.24}}    &\thead{\RS{14.55$\pm$0.07} \\ \RS{13.66$\pm$0.14} \\\textbf{\RS{24.96$\pm$0.31}}}    &\thead{\textbf{\RS{14.12$\pm$0.17}} \\ \textbf{\RS{13.54$\pm$0.33}} \\\RS{25.87$\pm$0.64}} \\ \midrule
    PEMS04       &\thead{MAE \\ MAPE (\%) \\RMSE}  &\thead{\RS{26.37$\pm$0.08} \\ \RS{19.89$\pm$1.10} \\\RS{40.19$\pm$0.12}}    &\thead{\RS{27.18$\pm$0.23} \\ \RS{21.94$\pm$1.46} \\\RS{40.95$\pm$0.25}}    &\thead{\RS{26.52$\pm$0.18} \\ \RS{20.17$\pm$1.67} \\\RS{40.19$\pm$0.03}}    &\thead{\RS{23.65$\pm$0.04} \\ \RS{16.05$\pm$0.10} \\\RS{37.12$\pm$0.07}}    &\thead{\RS{22.27$\pm$0.18} \\ \RS{14.36$\pm$0.12} \\\RS{35.02$\pm$0.19}}    &\thead{\RS{22.42$\pm$0.19} \\ \RS{15.87$\pm$0.36} \\\RS{34.75$\pm$0.19}}    &\thead{\RS{19.36$\pm$0.02} \\ \RS{13.31$\pm$0.19} \\\RS{31.72$\pm$0.13}}    &\thead{\RS{19.33$\pm$0.13} \\ \RS{12.83$\pm$0.07} \\\RS{31.23$\pm$0.21}}    &\thead{\RS{21.19$\pm$0.10} \\ \RS{13.90$\pm$0.05} \\\RS{33.65$\pm$0.20}}    &\thead{\RS{20.22$\pm$0.12} \\ \RS{13.62$\pm$0.02} \\\RS{31.97$\pm$0.17}}    &\thead{\RS{18.44$\pm$0.08} \\ \RS{12.37$\pm$0.08} \\\RS{31.02$\pm$0.18}}             &\textbf{\thead{\RS{18.11$\pm$0.02} \\ \RS{11.96$\pm$0.04} \\\RS{29.88$\pm$0.04}}} \\ \midrule
    PEMS07       &\thead{MAE \\ MAPE (\%) \\RMSE}  &\thead{\RS{29.56$\pm$0.22} \\ \RS{14.57$\pm$1.68} \\\RS{44.39$\pm$0.19}}    &\thead{\RS{30.59$\pm$0.25} \\ \RS{15.69$\pm$1.16} \\\RS{45.36$\pm$0.23}}    &\thead{\RS{29.31$\pm$0.16} \\ \RS{13.73$\pm$0.48} \\\RS{44.05$\pm$0.14}}    &\thead{\RS{23.60$\pm$0.05} \\ \RS{10.28$\pm$0.02} \\\RS{36.51$\pm$0.05}}    &\thead{\RS{27.41$\pm$0.45} \\ \RS{12.23$\pm$0.38} \\\RS{41.02$\pm$0.58}}    &\thead{\RS{25.98$\pm$0.78} \\ \RS{11.84$\pm$0.69} \\\RS{39.65$\pm$0.89}}    &\thead{\RS{21.22$\pm$0.24} \\ \RS{9.07$\pm$0.20} \\\RS{34.12$\pm$0.18}}     &\thead{\RS{20.73$\pm$0.08} \\ \RS{8.86$\pm$0.10} \\\RS{34.38$\pm$0.15}}     &\thead{\RS{24.26$\pm$0.14} \\ \RS{10.21$\pm$0.05} \\\RS{39.03$\pm$0.27}}    &\thead{\RS{20.61$\pm$0.14} \\ \RS{8.90$\pm$0.11} \\\RS{33.48$\pm$0.16}}    &\thead{\RS{19.26$\pm$0.17} \\ \RS{8.54$\pm$0.19} \\\RS{32.75$\pm$0.25}}              &\textbf{\thead{\RS{17.98$\pm$0.03} \\ \RS{7.50$\pm$0.02} \\\RS{31.20$\pm$0.04}}} \\ \midrule
    PEMS08       &\thead{MAE \\ MAPE (\%) \\RMSE}  &\thead{\RS{21.35$\pm$0.24} \\ \RS{15.71$\pm$1.77} \\\RS{32.45$\pm$0.11}}    &\thead{\RS{21.93$\pm$0.22} \\ \RS{17.11$\pm$2.07} \\\RS{33.14$\pm$0.14}}    &\thead{\RS{21.17$\pm$0.11} \\ \RS{13.97$\pm$0.53} \\\RS{32.31$\pm$0.08}}    &\thead{\RS{18.22$\pm$0.06} \\ \RS{11.56$\pm$0.04} \\\RS{28.29$\pm$0.09}}    &\thead{\RS{18.04$\pm$0.19} \\ \RS{11.16$\pm$0.10} \\\RS{27.94$\pm$0.18}}    &\thead{\RS{18.86$\pm$0.41} \\ \RS{12.50$\pm$0.66} \\\RS{28.55$\pm$0.49}}    &\thead{\RS{15.07$\pm$0.17} \\ \RS{9.51$\pm$0.22} \\\RS{23.85$\pm$0.18}}     &\thead{\RS{15.90$\pm$0.14} \\ \RS{10.53$\pm$0.21} \\\RS{25.00$\pm$0.17}}    &\thead{\RS{17.13$\pm$0.09} \\ \RS{10.96$\pm$0.07} \\\RS{26.18$\pm$0.18}}    &\thead{\RS{15.86$\pm$0.18} \\ \RS{10.50$\pm$0.20} \\\RS{24.98$\pm$0.15}}    &\thead{\RS{12.72$\pm$0.09} \\ \RS{8.78$\pm$0.20} \\\RS{22.60$\pm$0.13}}              &\textbf{\thead{\RS{12.54$\pm$0.01} \\ \RS{8.43$\pm$0.02} \\\RS{21.95$\pm$0.02}}} \\ 
    \bottomrule
    \end{tabular}
    }
    % }
    \caption{Performance comparison of different approaches for traffic flow forecasting.}
    \label{all_results}
\end{table*}

\begin{figure*}[t]
    \centering
    \subfloat[MAE]{%
        \includegraphics[width=0.3\textwidth]{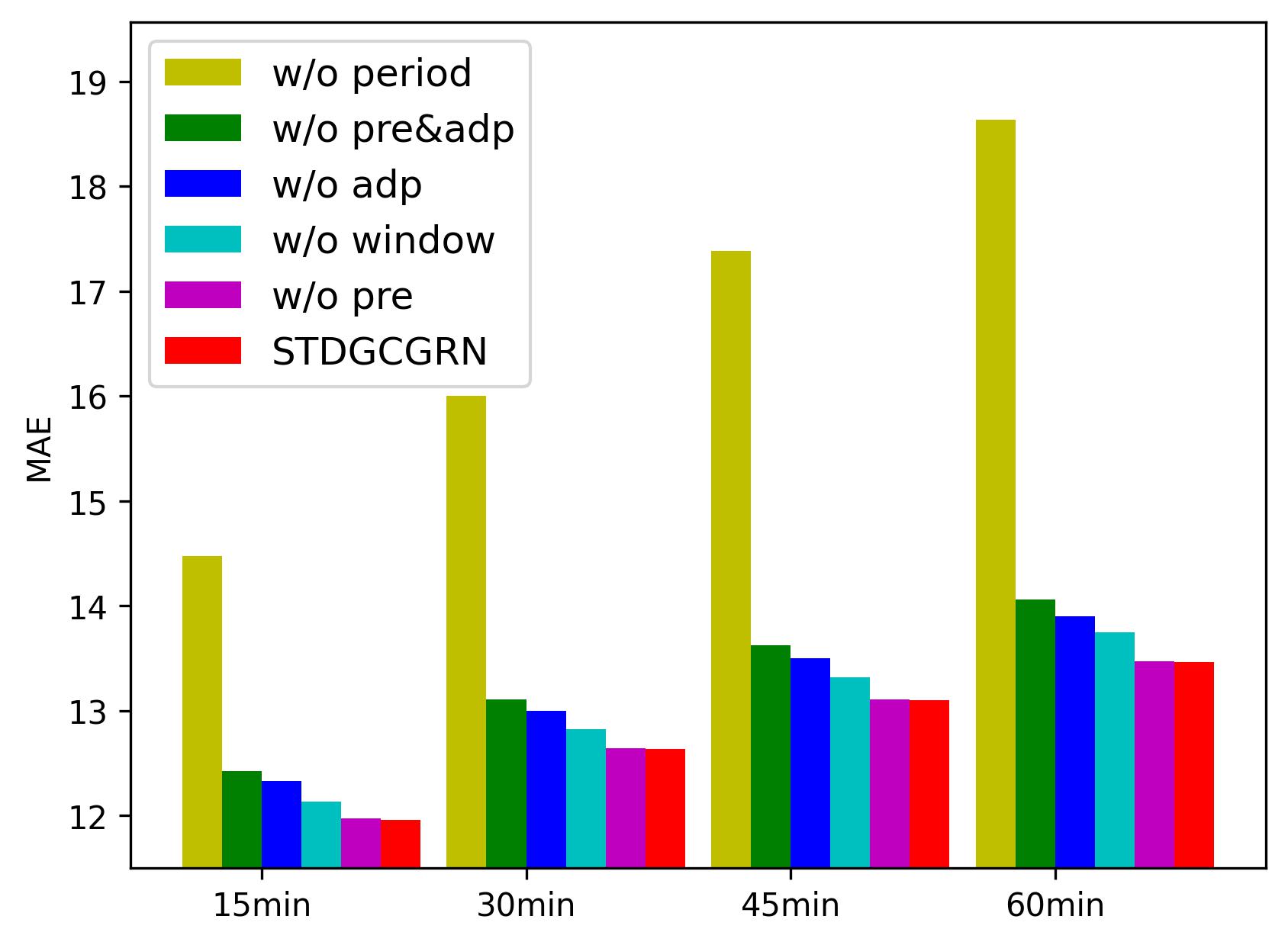}
        %\label{a_mae}
    }
    \subfloat[MAPE]{%
        \includegraphics[width=0.3\textwidth]{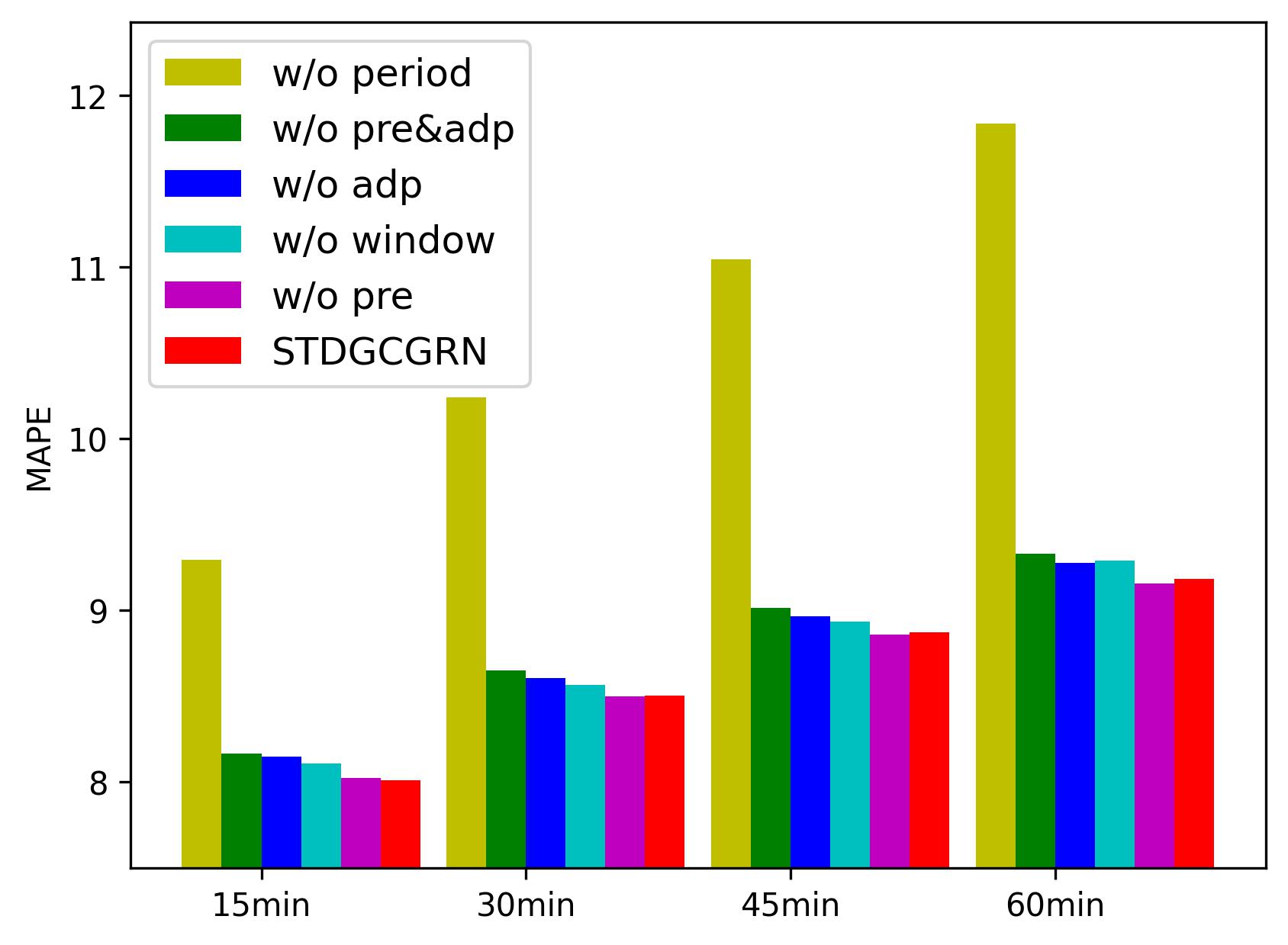}
        %\label{a_mape}
    } 
    \subfloat[RMSE]{%
        \includegraphics[width=0.3\textwidth]{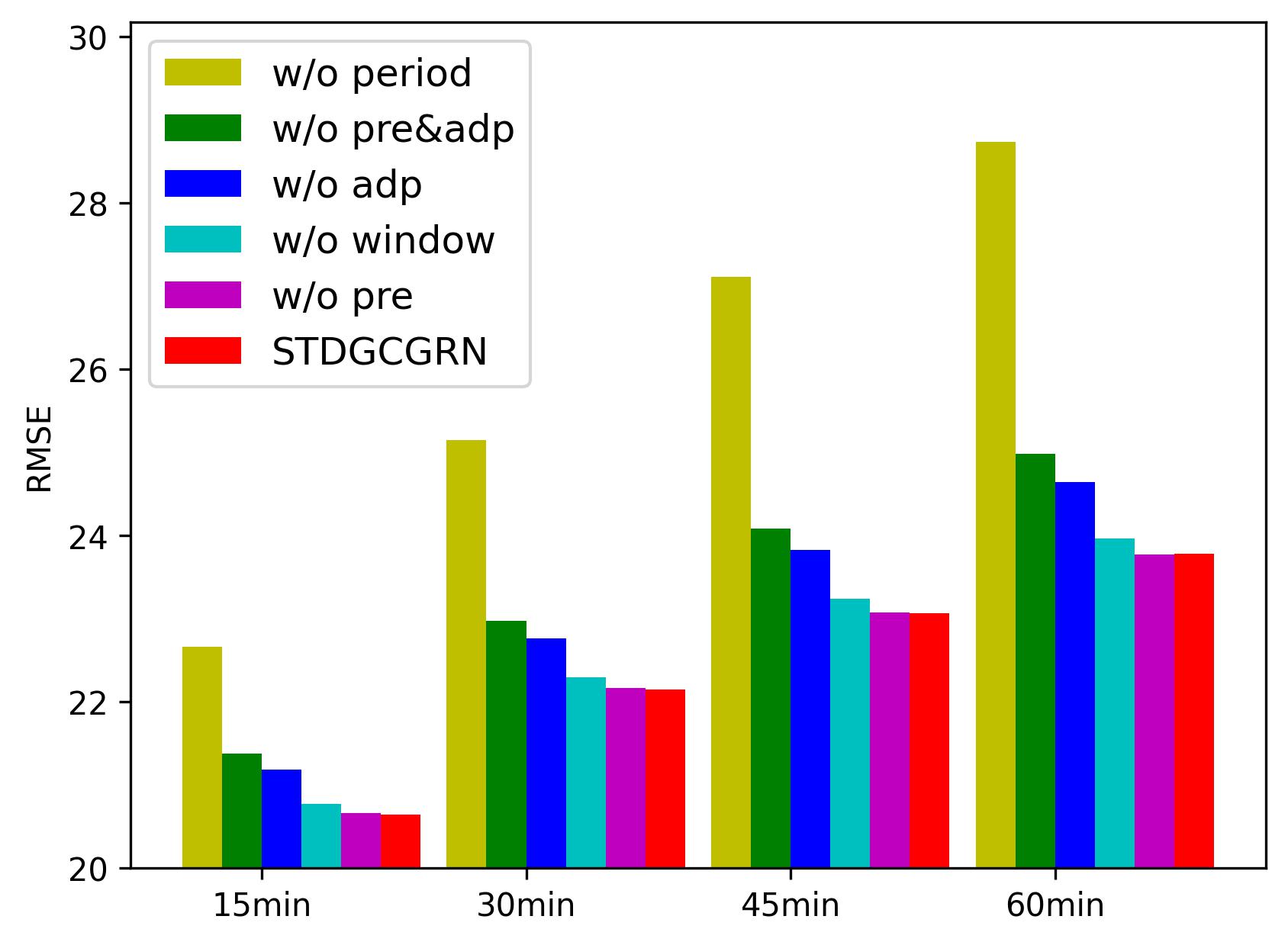}
        %\label{a_rmse}
    }
    \caption{Ablation study on the PEMS08 dataset.}
    \label{ablation_res}
\end{figure*}

\subsection{Ablation~Study} \label{ablation_study}
To further verify the effectiveness of different components of STGCGRN, we conduct ablation experiments on the PEMS08 dataset.
We design five variants of the STGCGRN model as follows:
\begin{itemize}
    \item w/o pre: We remove the distance-based predefined adjacency matrix from the DGCGRU layer.
    \item w/o adp: We remove the self-adaptive adjacency matrix from the DGCGRU layer.
    \item w/o pre\&adp: We remove the distance-based predefined adjacency matrix and self-adaptive adjacency matrix from the DGCGRU layer, replacing them with identity matrix.
    \item w/o window: We remove the sliding window in the attention module.
    \item w/o period: We remove the period information from STGCGRN.
\end{itemize}
We visualized the MAE, MAPE, RMSE of STGCGRN and its variants in the next 15min, 30min, 45min, 60min, as shown in the Figure~\ref{ablation_res}.
The introduction of periodic information has the most obvious improvement in model performance, which shows that long-term temporal dependency are critical to traffic forecasting.
In addition, with the growth of time, the effect of periodic information becomes more and more obvious.
This is because the error of the decoder gradually accumulate, and the period information can provide the decoder with historical data as a reference to help the decoder correct the error.
The effect of the hidden spatial dependency is second only to the period information.
%, while the improvement of the model by the distance-based spatial dependency is the smallest.
This is because the self-adaptive adjacency matrix is trained with the model, and the learned hidden dependency are directly related to the downstream task.%, i.e. traffic prediction. 
The predefined adjacency matrix can help to improve the performance of the model without the self-adaptive adjacency matrix, otherwise it has almost no effect, which indicates that the self-adaptive adjacency matrix has learned the part of the distance information that is helpful for prediction during the training process.
% In addition, the sliding window also helps to improve the performance of the model, indicating that the prior information that the time near the prediction time is of great help to the prediction task is important, which helps the attention mechanism to converge to a better position.
In addition, the sliding window also helps to improve the performance of the model, indicating that the time near the prediction time is of great help to the prediction task, which helps the attention mechanism to converge to a better position.

\subsection{Multi-Head~Mechanism~Study} \label{study_multihead}
\begin{table}
    \centering
    %\resizebox{\linewidth}{!}{
    \begin{tabular}{lccc} \toprule
                & MAE                       & MAPE                      &  RMSE\\ \midrule % \makecell{Multi-Head\\Co-Training} 
        1H      & \RS{12.62$\pm$0.04}       & \RS{8.46$\pm$0.05}        & \RS{22.13$\pm$0.04}\\
        2H      & \RS{12.58$\pm$0.02}       & \RS{8.45$\pm$0.02}        & \RS{22.08$\pm$0.01}\\
        4H      & \RS{12.56$\pm$0.01}       & \RS{8.44$\pm$0.04}       & \RS{22.02$\pm$0.03}\\
        8H      & \textbf{12.54$\pm$0.02}   & \textbf{8.43$\pm$0.02}   & \textbf{21.95$\pm$0.02} \\
        16H     & \RS{12.55$\pm$0.01}       & \RS{8.44$\pm$0.05}       & \RS{21.99$\pm$0.01} \\
        \bottomrule
    \end{tabular}
    %}
    \caption{Multi-Head mechanism study on the PEMS08 dataset.}
    \label{multihead}
\end{table}
Since there may be multiple hidden spatial dependencies between nodes, our work introduces the multi-head mechanism, hoping that $E_1$ and $E_2$ can model different hidden spatial dependencies. 
We conducted experiments to verify that the introduction of the multi-head mechanism improves the performance of the model. 
The experimental results are shown in the Table~\ref{multihead}.
Increasing the number of heads bring performance gain.
The model performs best when the number of heads is 8. 
This shows that as the number of heads increases, the model captures more hidden spatial dependency information, which verify the effectiveness of the multi-head mechanism in the current scenario.
%When the number of heads reaches 16, the performance of the model decreases slightly. This may be because the dimension of the vector representation of each head is only 8 when the number of heads is 18 (we set the sum of the dimensions of all heads to 128), which cannot well represent the node.
\subsection{Position of DGCGRU layer}
\begin{table}
    \centering
    \resizebox{\linewidth}{!}{
    \begin{tabular}{lccc} \toprule
                          & MAE                       & MAPE                      &  RMSE\\ \midrule % \makecell{Multi-Head\\Co-Training} 
        STGCGRN          & \RS{12.80$\pm$0.01}       & \RS{8.60$\pm$0.03}        & \RS{22.17$\pm$0.02}\\
        STGCGRN\_rev      & \textbf{12.54$\pm$0.02}   & \textbf{8.43$\pm$0.02}   & \textbf{21.95$\pm$0.02} \\
        \bottomrule
    \end{tabular}
    }
    \caption{Comparative experiment on the PEMS08 dataset.}
    \label{comparative_exp}
\end{table}
We explore how to better combine single-node periodic information and multi-node spatial dependencies to prevent periodic information of nodes from interfering with each other.
The result is shown in Table \ref{comparative_exp}. STGCGRN\_rev is a variant of STGCGRN where the order of the DGCGRU layer and the attention module is reversed.

We can see that the performance of STGCGRN\_rev decline, indicating that modeling spatial dependencies first will cause the periodic information of each node to interfere with each other when mining periodic information later. The architecture of STGCGRN is better.

\section{Conclusion}
In this paper, we proposed Spatial-Temporal Graph Convolutional Gated Recurrent Network (STGCGRN) for traffic forecasting.
For temporal dependency, we model short-term dependency by processing recent traffic data with GRU, and capture long-term dependency by mining daily and weekly periodicity. 
For spatial dependency, we model distance-based spatial dependency and multiple hidden spatial dependencies with predefined adjacency matrix and multi-head self-adaptive adjacency matrix, respectively.
In addition, we also explored the model structure to avoid the mutual interference of periodic information among nodes during spatial dependency modeling.
Experiments and analysis on four datasets show that our model achieves state-of-the-art results.
The code have been released at: https://github.com/ZLBryant/STGCGRN.

\bibliography{ref}

\begin{thebibliography}{35}
\providecommand{\natexlab}[1]{#1}

\bibitem[{Ahmed and Cook(1979)}]{Ahmed1979ANALYSISOF}
Ahmed, M.~S.; and Cook, A.~R. 1979.
\newblock Analysis of freeway traffic time-series data by using box-jenkins
  techniques.
\newblock \emph{Transportation Research Record}, (722): 1--9.

\bibitem[{Atwood and Towsley(2016)}]{Atwood2016DiffusionConvolutionalNN}
Atwood, J.; and Towsley, D.~F. 2016.
\newblock Diffusion-Convolutional Neural Networks.
\newblock In \emph{In Advances in Neural Information Processing Systems},
  1993--2001.

\bibitem[{Bai et~al.(2020)Bai, Yao, Li, Wang, and Wang}]{Bai2020AdaptiveGC}
Bai, L.; Yao, L.; Li, C.; Wang, X.; and Wang, C. 2020.
\newblock Adaptive Graph Convolutional Recurrent Network for Traffic
  Forecasting.
\newblock In \emph{In Advances in Neural Information Processing Systems},
  17804--17815.

\bibitem[{Bruna et~al.(2014)Bruna, Zaremba, Szlam, and
  LeCun}]{Bruna2014SpectralNA}
Bruna, J.; Zaremba, W.; Szlam, A.~D.; and LeCun, Y. 2014.
\newblock Spectral Networks and Locally Connected Networks on Graphs.
\newblock In \emph{Proceedings of the International Conference on Learning
  Representations}, 1--14.

\bibitem[{Chen et~al.(2001)Chen, Petty, Skabardonis, Varaiya, and
  Jia}]{chen2001freeway}
Chen, C.; Petty, K.; Skabardonis, A.; Varaiya, P.; and Jia, Z. 2001.
\newblock Freeway performance measurement system: mining loop detector data.
\newblock \emph{Transportation Research Record}, 1748(1): 96--102.

\bibitem[{Davis and Nihan(1991)}]{Davis1991NonparametricRA}
Davis, G.~A.; and Nihan, N.~L. 1991.
\newblock Nonparametric Regression and Short Term Freeway Traffic Forecasting.
\newblock \emph{Journal of Transportation Engineering-asce}, 117(2): 178--188.

\bibitem[{Defferrard, Bresson, and
  Vandergheynst(2016)}]{Defferrard2016ConvolutionalNN}
Defferrard, M.; Bresson, X.; and Vandergheynst, P. 2016.
\newblock Convolutional Neural Networks on Graphs with Fast Localized Spectral
  Filtering.
\newblock In \emph{In Advances in Neural Information Processing Systems,},
  3844--3852.

\bibitem[{Fang et~al.(2019)Fang, Zhang, Meng, Xiang, and
  Pan}]{Fang2019GSTNetGS}
Fang, S.; Zhang, Q.; Meng, G.; Xiang, S.; and Pan, C. 2019.
\newblock GSTNet: Global spatial-temporal network for traffic flow prediction.
\newblock In \emph{International Joint Conference on Artificial Intelligence},
  2286--2293.

\bibitem[{Fu, Zhang, and Li(2016)}]{fu2016using}
Fu, R.; Zhang, Z.; and Li, L. 2016.
\newblock Using LSTM and GRU neural network methods for traffic flow
  prediction.
\newblock In \emph{Youth Academic Annual Conference of Chinese Association of
  Automation}, 324--328.

\bibitem[{Gilmer et~al.(2017)Gilmer, Schoenholz, Riley, Vinyals, and
  Dahl}]{gilmer2017neural}
Gilmer, J.; Schoenholz, S.~S.; Riley, P.~F.; Vinyals, O.; and Dahl, G.~E. 2017.
\newblock Neural message passing for quantum chemistry.
\newblock In \emph{International conference on machine learning}, 1263--1272.

\bibitem[{Guo et~al.(2019)Guo, Lin, Feng, Song, and Wan}]{Guo2019AttentionBS}
Guo, S.; Lin, Y.; Feng, N.; Song, C.; and Wan, H. 2019.
\newblock Attention Based Spatial-Temporal Graph Convolutional Networks for
  Traffic Flow Forecasting.
\newblock In \emph{Proceedings of the AAAI conference on artificial
  intelligence}, 922--929.

\bibitem[{Guo et~al.(2021)Guo, Lin, Wan, Li, and Cong}]{guo2021learning}
Guo, S.; Lin, Y.; Wan, H.; Li, X.; and Cong, G. 2021.
\newblock Learning dynamics and heterogeneity of spatial-temporal graph data
  for traffic forecasting.
\newblock \emph{IEEE Transactions on Knowledge and Data Engineering}.

\bibitem[{Hamilton, Ying, and Leskovec(2017)}]{Hamilton2017InductiveRL}
Hamilton, W.~L.; Ying, Z.; and Leskovec, J. 2017.
\newblock Inductive Representation Learning on Large Graphs.
\newblock In \emph{In Advances in Neural Information Processing Systems},
  1024--1034.

\bibitem[{Kipf and Welling(2017)}]{Kipf2017SemiSupervisedCW}
Kipf, T.; and Welling, M. 2017.
\newblock Semi-Supervised Classification with Graph Convolutional Networks.
\newblock In \emph{Proceedings of the International Conference on Learning
  Representations}, 1--14.

\bibitem[{Lee and b.~Fambro(1999)}]{Lee1999ApplicationOS}
Lee, S.; and b.~Fambro, D. 1999.
\newblock Application of Subset Autoregressive Integrated Moving Average Model
  for Short-Term Freeway Traffic Volume Forecasting.
\newblock \emph{Transportation Research Record}, 1678(1): 179--188.

\bibitem[{Li et~al.(2021{\natexlab{a}})Li, Feng, Yan, Jin, Jin, and
  Li}]{li2021dynamic}
Li, F.; Feng, J.; Yan, H.; Jin, G.; Jin, D.; and Li, Y. 2021{\natexlab{a}}.
\newblock Dynamic graph convolutional recurrent network for traffic prediction:
  Benchmark and solution.
\newblock \emph{arXiv preprint arXiv:2104.14917}.

\bibitem[{Li et~al.(2021{\natexlab{b}})Li, Feng, Yan, Jin, Jin, and
  Li}]{Li2021DynamicGC}
Li, F.; Feng, J.; Yan, H.; Jin, G.; Jin, D.; and Li, Y. 2021{\natexlab{b}}.
\newblock Dynamic Graph Convolutional Recurrent Network for Traffic Prediction:
  Benchmark and Solution.
\newblock \emph{ArXiv}, abs/2104.14917.

\bibitem[{Li et~al.(2018)Li, Yu, Shahabi, and Liu}]{Li2018DiffusionCR}
Li, Y.; Yu, R.; Shahabi, C.; and Liu, Y. 2018.
\newblock Diffusion Convolutional Recurrent Neural Network: Data-Driven Traffic
  Forecasting.
\newblock In \emph{Proceedings of the International Conference on Learning
  Representations}, 1--16.

\bibitem[{Lin et~al.(2019)Lin, Feng, Lu, Li, and Jin}]{lin2019deepstn+}
Lin, Z.; Feng, J.; Lu, Z.; Li, Y.; and Jin, D. 2019.
\newblock Deepstn+: Context-aware spatial-temporal neural network for crowd
  flow prediction in metropolis.
\newblock In \emph{Proceedings of the AAAI conference on artificial
  intelligence}, volume~33, 1020--1027.

\bibitem[{Luong, Pham, and Manning(2015)}]{Luong2015EffectiveAT}
Luong, T.; Pham, H.; and Manning, C.~D. 2015.
\newblock Effective Approaches to Attention-based Neural Machine Translation.
\newblock In \emph{Proceedings of the 2015 Conference on Empirical Methods in
  Natural Language Processing}, 1412--1421.

\bibitem[{Shuman et~al.(2013)Shuman, Narang, Frossard, Ortega, and
  Vandergheynst}]{Shuman2013TheEF}
Shuman, D.~I.; Narang, S.~K.; Frossard, P.; Ortega, A.; and Vandergheynst, P.
  2013.
\newblock The emerging field of signal processing on graphs: Extending
  high-dimensional data analysis to networks and other irregular domains.
\newblock \emph{IEEE signal processing magazine}, 30(3): 83--98.

\bibitem[{Song et~al.(2020)Song, Lin, Guo, and Wan}]{song2020spatial}
Song, C.; Lin, Y.; Guo, S.; and Wan, H. 2020.
\newblock Spatial-temporal synchronous graph convolutional networks: A new
  framework for spatial-temporal network data forecasting.
\newblock In \emph{Proceedings of the AAAI Conference on Artificial
  Intelligence}, volume~34, 914--921.

\bibitem[{Van Der~Voort, Dougherty, and Watson(1996)}]{van1996combining}
Van Der~Voort, M.; Dougherty, M.; and Watson, S. 1996.
\newblock Combining Kohonen maps with ARIMA time series models to forecast
  traffic flow.
\newblock \emph{Transportation Research Part C: Emerging Technologies}, 4(5):
  307--318.

\bibitem[{Wang et~al.(2017)Wang, Pan, Long, Zhu, and Jiang}]{wang2017mgae}
Wang, C.; Pan, S.; Long, G.; Zhu, X.; and Jiang, J. 2017.
\newblock Mgae: Marginalized graph autoencoder for graph clustering.
\newblock In \emph{Proceedings of the 2017 ACM on Conference on Information and
  Knowledge Management}, 889--898.

\bibitem[{Wang et~al.(2020)Wang, Ma, Wang, Jin, Wang, Tang, Jia, and
  Yu}]{wang2020traffic}
Wang, X.; Ma, Y.; Wang, Y.; Jin, W.; Wang, X.; Tang, J.; Jia, C.; and Yu, J.
  2020.
\newblock Traffic flow prediction via spatial temporal graph neural network.
\newblock In \emph{Proceedings of The Web Conference 2020}, 1082--1092.

\bibitem[{Williams(2001)}]{Williams2001MultivariateVT}
Williams, B.~M. 2001.
\newblock Multivariate Vehicular Traffic Flow Prediction: Evaluation of ARIMAX
  Modeling.
\newblock \emph{Transportation Research Record}, 1776(1): 194--200.

\bibitem[{Wu, Ho, and Lee(2004)}]{Wu2004TraveltimePW}
Wu, C.-H.; Ho, J.-M.; and Lee, D.~T. 2004.
\newblock Travel-time prediction with support vector regression.
\newblock \emph{IEEE Transactions on Intelligent Transportation Systems}, 5(4):
  276--281.

\bibitem[{Wu et~al.(2019)Wu, Pan, Long, Jiang, and Zhang}]{Wu2019GraphWF}
Wu, Z.; Pan, S.; Long, G.; Jiang, J.; and Zhang, C. 2019.
\newblock Graph WaveNet for Deep Spatial-Temporal Graph Modeling.
\newblock In \emph{International Joint Conference on Artificial Intelligence},
  1907--1913.

\bibitem[{Yao et~al.(2019{\natexlab{a}})Yao, Tang, Wei, Zheng, and
  Li}]{yao2019revisiting}
Yao, H.; Tang, X.; Wei, H.; Zheng, G.; and Li, Z. 2019{\natexlab{a}}.
\newblock Revisiting spatial-temporal similarity: A deep learning framework for
  traffic prediction.
\newblock In \emph{Proceedings of the AAAI conference on artificial
  intelligence}, volume~33, 5668--5675.

\bibitem[{Yao et~al.(2019{\natexlab{b}})Yao, Tang, Wei, Zheng, and
  Li}]{Yao2019RevisitingSS}
Yao, H.; Tang, X.; Wei, H.; Zheng, G.; and Li, Z. 2019{\natexlab{b}}.
\newblock Revisiting spatial-temporal similarity: A deep learning framework for
  traffic prediction.
\newblock In \emph{Proceedings of the AAAI conference on artificial
  intelligence}, 5668--5675.

\bibitem[{Ying et~al.(2018)Ying, You, Morris, Ren, Hamilton, and
  Leskovec}]{ying2018hierarchical}
Ying, Z.; You, J.; Morris, C.; Ren, X.; Hamilton, W.; and Leskovec, J. 2018.
\newblock Hierarchical graph representation learning with differentiable
  pooling.
\newblock \emph{Advances in neural information processing systems}, 31.

\bibitem[{Yu, Yin, and Zhu(2018)}]{Yu2018SpatioTemporalGC}
Yu, B.; Yin, H.; and Zhu, Z. 2018.
\newblock Spatio-Temporal Graph Convolutional Networks: A Deep Learning
  Framework for Traffic Forecasting.
\newblock In \emph{Proceedings of the International Joint Conference on
  Artificial Intelligence}, 3634--3640.

\bibitem[{Yu et~al.(2017)Yu, Li, Shahabi, Demiryurek, and Liu}]{yu2017deep}
Yu, R.; Li, Y.; Shahabi, C.; Demiryurek, U.; and Liu, Y. 2017.
\newblock Deep learning: A generic approach for extreme condition traffic
  forecasting.
\newblock In \emph{Proceedings of the 2017 SIAM international Conference on
  Data Mining}, 777--785. SIAM.

\bibitem[{Zhang, Zheng, and Qi(2017)}]{zhang2017deep}
Zhang, J.; Zheng, Y.; and Qi, D. 2017.
\newblock Deep spatio-temporal residual networks for citywide crowd flows
  prediction.
\newblock In \emph{Thirty-first AAAI conference on artificial intelligence},
  1655--1661.

\bibitem[{Zhang and Chen(2018)}]{zhang2018link}
Zhang, M.; and Chen, Y. 2018.
\newblock Link prediction based on graph neural networks.
\newblock \emph{Advances in neural information processing systems}, 31.

\end{thebibliography}
\end{document}